# IMPROVING DISTRIBUTION ALIGNMENT WITH DIVERSITY-BASED SAMPLING

*Andrea Napoli, Paul White*

Institute of Sound and Vibration Research
University of Southampton, UK
{an1g18, P.R.White}@soton.ac.uk

**ABSTRACT**

Domain shifts are ubiquitous in machine learning, and can substantially degrade a model's performance when deployed to real-world data. To address this, distribution alignment methods aim to learn feature representations which are invariant across domains, by minimising the discrepancy between the distributions. However, the discrepancy estimates can be extremely noisy when training via stochastic gradient descent (SGD), and shifts in the relative proportions of different subgroups can lead to domain misalignments; these can both stifle the benefits of the method. This paper proposes to improve these estimates by inducing diversity in each sampled minibatch. This simultaneously balances the data and reduces the variance of the gradients, thereby enhancing the model's generalisation ability. We describe two options for diversity-based data samplers, based on the $k$-determinantal point process ($k$-DPP) and the $k$-means++ algorithm, which can function as drop-in replacements for a standard random sampler. On a real-world domain shift task of bioacoustic event detection, we show that both options 1) yield minibatches which are more representative of the full dataset; 2) reduce the distance estimation error between distributions, for a given sample size; and 3) improve out-of-distribution accuracy for two distribution alignment algorithms, as well as standard ERM.

## 1. INTRODUCTION

Machine learning methods often underperform on data lying outside the training distribution. The sensitivity to distributional shifts (also called domain shifts) is currently a severe limitation to the widespread deployment of AI to real-world problems. This issue manifests widely, and significant research effort has already been invested towards achieving better out of distribution (OOD) generalisation [1, 2].

Distribution alignment (also referred to as invariant feature learning or invariance regularisation) is possibly the dominant approach in this field. Given meta-data which groups training examples according to certain characteristics or contexts (referred to as *domains*), the technique aims to learn feature representations which are invariant to these characteristics, such that the model becomes robust to future changes in the data. If unlabelled data from the test domain is included, this technique is referred to as unsupervised domain adaptation (UDA).

In practice, distribution alignment has manifested in two main ways:

1) as a direct (differentiable) measure of the distance between the statistics of different distributions, which is minimised alongside the standard objective of empirical risk (ERM). Such statistics include the distributions' means or covariances [3], or the kernel mean embeddings [4].

2) via domain-adversarial training [5, 6], in which a discriminator network is trained to predict which domain the features belong to, and the feature extractor is tuned to maximise discriminator error, alongside the loss on the main task. Depending on the exact formulation, it has been shown that adversarial networks minimise the Jensen-Shannon divergence or Wasserstein distance between domains [7].

Although these techniques employ clever tricks that bypass the need for direct density estimation, the distributions must still be properly characterised by the samples if they are to be properly aligned. In this context, where the distributions are complex, the feature space is high-dimensional and the sample sizes are small, distribution coverage is inevitably poor. In practice, and possibly because of this, distribution alignment has frequently been found to have a negligible or even negative impact on training compared to vanilla ERM [1, 2, 8–10].

In this paper, we propose that minibatches that better cover the support of their underlying distribution would give higher quality distance estimates, and thus increase the effectiveness of alignment methods. We propose to achieve this by inducing diversity in each sampled minibatch – corresponding to the datapoints being "spread out" (pairwise dissimilar) in the learned model's feature space.

We note that this approach can also be interpreted as a generalisation of class-balancing [11]. Inevitably, complex real-world acoustic scenes have a far richer ontology than the fixed set of class labels provided for the specific learning task (which may only be binary). So, this method can be motivated by the same logic as why classes are normally balanced prior to training: to ensure equal representation of all sound events.

Thus, the requirement is for a fast, scalable sampler which can stochastically draw independent, diverse minibatches of fixed cardinality from the corpus. It should also be possible to weight each instance to bias its selection probability based on prior knowledge, e.g., the label distribution – although, note, we are not interested in explicitly class-balancing the data, as doing so is at odds with the objective of diversity: some classes (e.g., "not a humpback whale") may have far greater variety than others. Also note that, given feature-label continuity, inducing diversity does tend to implicitly class-balance the data anyway [12].

We identify two options which satisfy these desiderata: the $k$-determinantal point process ($k$-DPP) and the $k$-means++ algorithm, which are discussed next.

### 1.1. Determinantal point process (DPP)

Given a set of feature embeddings $\mathcal{X} = \{x_1, ..., x_n\}$, $x_i \in \mathbb{R}^d$, a point process on $\mathcal{X}$ is a probability measure over "point configurations" (i.e., subsets) of $\mathcal{X}$. Sampling a point process is thus equivalent to randomly drawing a subset of $\mathcal{X}$. For a determinantal point process (DPP) [13], the probability of drawing subset $\mathcal{A}$ is proportional to the determinant of a likelihood kernel $L_\mathcal{A}$ describing pairwise similarities between its elements. Specifically:

$$\mathbb{P}[\mathcal{A}] = \frac{\det L_\mathcal{A}}{\det [I + L]}, \qquad \forall \mathcal{A} \subseteq \mathcal{X}, \qquad (1)$$

where $L \in \mathbb{R}^{n \times n}$ is the kernel over all $\mathcal{X}$. When the DPP is conditioned to a fixed cardinality $|\mathcal{A}| = k \leq \mathrm{rank}(L)$, this is known as a $k$-DPP.

As a result of (1), large off-diagonal entries in $L$ imply low probability of co-occurrence in $\mathcal{A}$; this makes DPPs a common tool for inducing diversity. Previous use cases of DPPs in bioacoustics include to facilitate the exploration of large corpuses [14], and to implicitly class-balance unlabelled data for semi-supervised learning and unsupervised domain adaptation [12]. DPPs have also been applied as a variance reduction and balancing scheme for minibatch gradient descent, but without specific application to distribution alignment techniques [11].

To apply weights $w = [w_1, ..., w_n]^T$ to each instance, we can define $L$ based on a similarity matrix $S$, with each element weighted by the corresponding pair of weights: $L_{ij} = \sqrt{w_i w_j} S_{ij}$.

Thus, the $k$-DPP is now restricted to $k \leq \mathrm{rank}(S)$. An appropriate choice of similarity measure should ensure that $S$ is full rank (that is, the kernel should be *strictly* positive-definite), so as not to limit the minibatch size we can use. For example, the commonly-used linear kernel $S_{ij} = x_i^T x_j$ results in *at best* $k \leq d$, but this could well be lower if the features are not all linearly independent. Therefore, in this paper, we propose to use the radial basis function (RBF) kernel instead. Specifically, adopting a common heuristic for the bandwidth parameter $\gamma$ [4], we use an RBF mixture kernel defined by

$$S_{ij} = \sum_{\gamma \in \mathcal{G}} e^{-\gamma \|x_i - x_j\|^2} \qquad (2)$$

with $\mathcal{G} = \{0.001, 0.01, 0.1, 1, 10\}$. That the RBF is always strictly positive-definite is a well-known result [15].

### 1.2. $k$-means++

$k$-means++ [16] is an algorithm originally envisioned as an initialisation for k-means clustering, designed to select a subset of highly dissimilar points from a corpus. Again, weights can easily be applied to each instance. The algorithm is as follows:

1) Choose an initial point at random from $\mathcal{X}$ with probabilities weighted by $w$. Remove the point from $\mathcal{X}$ and append to $\mathcal{A}$.

2) For each $x_i \in \mathcal{X}$, compute $D(x_i) = \min_{x' \in \mathcal{A}} \|x_i - x'\|$, the distance between $x_i$ and the closest point in $\mathcal{A}$.

3) Choose the next point with probability $\propto w_i D(x_i)^2$.

4) Repeat steps 2 and 3 until $k$ points are chosen.

### 1.3. Training strategy

Ideally, the samplers would have access to up-to-date feature embeddings for every draw. However, recomputing $\mathcal{X}$ (not to mention $S$) at every training iteration would be slow. Instead, we propose to only update the samplers periodically every $t$ iterations; $t$ is thus a trade-off between training speed and the quality of the similarity information in $S$. Where no pretrained feature extractor is available, the first $t$ iterations are performed with standard (weighted) random samplers, although we posit that using the diversity-based samplers with features from a newly-initialised network with random weights would have an equivalent effect.

## 2. EXPERIMENTS

In this section, we evaluate the proposed method on a real-world domain shift problem, namely, the detection of humpback whale calls across data from different acoustic monitoring programs [8]. The dataset comprises 43,385 samples split roughly equally across 4 recording locations (Madagascar, UK, Hawaii, and Australia). Each sample is a PCEN-normalised [17] mel-spectrogram of a 4-second audio clip sampled at 10 kHz, labelled as either "humpback whale" or "not humpback whale". Some exemplar spectrograms are shown in Figure 1 (note, these images are linear-scaled and pre-PCEN). A simple 4-layer CNN architecture is used as the core model, with 16 filters per layer and RELU activations.

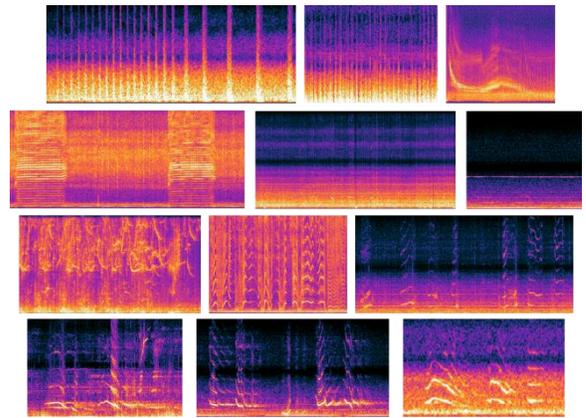

Figure 1: Some exemplar spectrograms of sounds in the dataset (5 kHz bandwidth, time axis scales variable). Top row: sperm whale clicks, pilot whale clicks, seal vocalisations. Second row: minke whale boings, right whale calls in strong vessel noise, electrical interference. Third row: dolphin whistles, dolphin creaks, right whale calls. Bottom row: three humpback whale calls.

Experiments are conducted using the DomainBed framework [1]. This means 3 locations ("domains") are used at a time for training and the remaining domain for testing. Models are trained for 2,000 iterations, with the samplers updated every 400. Hyperparameters are chosen via random search of size 40 using an oracle validation set (i.e., a set following the same distribution as the test set) as this provides the greatest stability for hyperparameter tuning and reduces the noise in the results from this source. Experiments are repeated 5 times for reproducibility, using different random seeds for hyperparameters, weight initialisations, and dataset splits. All other options follow the DomainBed defaults.

We use the DPPy Python package [18] for $k$-DPP sampling (specifically, the exact spectral sampler) and the scikit-learn [19] implementation of $k$-means++. The features used for sampling are the same features used by the alignment methods (i.e., the activations from the last convolutional layer of the model). See [8] for more model and dataset details and [1] for further training and hyperparameter details.

Table 1: Test domain accuracy (%) for each sampler and training algorithm.

|          | DG | | | UDA | | |
|----------|------|------|------|------|------|------|
| Sampler | ERM | CORAL | DANN | CORAL | DANN | Average |
| Random | 91.3 ± 0.7 | 86.2 ± 0.2 | 82.1 ± 1.2 | 90.4 ± 0.8 | 81.7 ± 1.6 | 86.3 ± 0.5 |
| $k$-DPP | 91.6 ± 0.4 | 90.6 ± 0.2 | **87.7 ± 0.9** | **94.0 ± 0.2** | 85.2 ± 1.1 | 89.8 ± 0.3 |
| $k$-means++ | **92.9 ± 0.7** | **91.5 ± 0.7** | 87.3 ± 1.6 | 93.8 ± 0.2 | **86.6 ± 1.7** | **90.4 ± 0.5** |

## 2.1. Impact on generalisation performance

First, we compare the effect of the samplers on the generalisation power of the trained models. We do this for 2 alignment algorithms: correlation alignment (CORAL) [3] and domain-adversarial neural networks (DANN) [5], in both adaptive (UDA) and non-adaptive (domain generalisation, DG) paradigms. In the DG setting, these are used only to align the 3 training domains to each other. For UDA, in addition to this, the training domains are also aligned to an unlabelled, held-out subset of the test domain (that is, *not* the same samples that are used to determine accuracy, nor tune hyperparameters). We also test ERM, which does not explicitly perform domain alignment and by its nature is DG only.

In addition to our 2 proposed diversity-based data samplers ($k$-DPP and $k$-means++), we compare a baseline of standard class-weighted random sampling. Our performance metric is average model accuracy across the 4 test domains, reported in Table 1, along with the standard error across the 5 repeats.

Firstly, our results reproduce findings from previous work [1, 2, 8–10]: with standard random samplers, both alignment methods perform poorly, underperforming ERM by as much as 10%. The results clearly show that using diversity-based sampling improves these methods, with consistent accuracy gains of 4 to 5 percentage points. Interestingly, ERM is also slightly improved, suggesting a general benefit to ensuring equal representation of all sound events.

Despite these gains, both CORAL and DANN still underperform ERM in the DG setting, showing just how difficult the DG problem is – as well as how strong the ERM baseline is. However, in the UDA setting, diversity-based sampling enables CORAL to finally exceed ERM, achieving the highest performance out of all the methods we test.

On average, accuracy is slightly higher with $k$-means++ than with the $k$-DPP, although this is within margin of uncertainty. In addition, $k$-means++ is computationally faster, easier to scale, and perhaps also more intuitive to understand and implement, making it the more favourable method overall.

So, we have shown that inducing diversity allows models trained with distribution alignment to generalise better to new domains. In Section 1, we claimed that this is because diverse samples are more representative of their underlying distributions, and that this reduces error when estimating the distances between distributions. We test both parts of this claim next.

## 2.2. Improved distribution coverage

Recall our aim is to choose subsets $\mathcal{A}$ which are more representative of the full set $\mathcal{X}$. This can be recognised as the problem of vector quantisation. Thus, a measure of the "representativeness" of $\mathcal{A}$ is a low value of the quantisation error (QE)

$$\text{QE} = \sum_{x_i \in \mathcal{X}} \min_{x' \in \mathcal{A}} \|x_i - x'\|^2, \qquad (3)$$

that is, the sum of squared distances between each $x_i$ and the closest point in $\mathcal{A}$.

Table 2 compares the average QE of the 3 samplers over 1000 independent draws of $\mathcal{A}$ from each domain, with $k = 32$, and based on the features extracted by the ERM models from Section 2.1. The QE of the $k$-DPP and $k$-means++ are shown to be greatly reduced compared to the random sampler, by 36% and 65% respectively. Given the direct connection between (3) and the $k$-means++ selection criterion, it is perhaps unsurprising that the QE is so much lower for the latter, although this has not translated into greater generalisation power to the same extent.

Table 2: Average quantisation error for each sampler.

| Sampler | QE |
|---------|-----|
| Random | 6861 ± 23 |
| $k$-DPP | 4418 ± 12 |
| $k$-means++ | **2425 ± 6** |

## 2.3. Lower-error distance estimates

Finally, we test the claim that diverse sampling improves distance estimation between distributions. To do this, we compare the estimation error of a popular distance estimate (the MMD) applied to the features of our multi-domain dataset.

Let $\mathcal{F} = \mathbb{R}^d$ be the feature space induced by our model. The MMD is computed on the basis of a positive-definite kernel $\kappa : \mathcal{F} \times \mathcal{F} \to \mathbb{R}$ and is defined as the distance between distribution means embedded in the reproducing kernel Hilbert space $\mathcal{H}$ associated with $\kappa$. For 2 distributions $\mathbb{P}_1, \mathbb{P}_2 \in \mathcal{P}$, we have

$$\text{MMD}(\mathbb{P}_1, \mathbb{P}_2) = \|\mu(\mathbb{P}_1) - \mu(\mathbb{P}_2)\|_\mathcal{H}, \qquad (4)$$

where $\mu : \mathcal{P} \to \mathcal{H}$ is the mean map operation

$$\mu(\mathbb{P}) = \mathbb{E}_{X \sim \mathbb{P}}[\phi(X)] \cong \frac{1}{n} \sum_{i=1}^{n} \phi(x_i) \qquad (5)$$

and $\phi : \mathcal{F} \to \mathcal{H}$ is the implicit mapping associated with $\mathcal{H}$. It has been shown that for certain *characteristic* kernels, including the RBF, $\mu$ is injective, meaning every possible feature distribution $\mathbb{P} \in \mathcal{P}$ is uniquely represented in $\mathcal{H}$ and the MMD is 0 if and only if the distributions are identical [20].

Concretely, the method is as follows. We train a model by ERM on 3 domains at a time, as in Section 2.1. Our target is to compute the average pairwise MMD between these 3 domains, based on features extracted from the model and the same RBF mixture kernel defined in (2). We compute a set of 1000 MMDs using only 32 examples per domain, drawn stochastically using each of the 3 samplers. We measure the error of these w.r.t. a "ground-truth" MMD computed using all the available data (~8000 examples per domain). We can do this because (5) is a consistent estimator of the embedded distribution mean: thus, an estimate computed with a larger sample will (in expectation) be closer to the true value of the MMD. Specifically, we compute the mean absolute percentage error (MAPE) in the MMDs, defined as

$$\text{MAPE} = 100\% \frac{1}{1000 D} \sum_{r=1}^{1000} |D - \widehat{D}_r| \quad (6)$$

where $D$ is the "ground-truth" MMD computed using the full dataset and $\widehat{D}_r$ are the MMDs computed using only 32 examples per domain. As before, we do this for all 4 combinations of training domains, and repeat 5 times for reproducibility. The results are shown in Table 3.

Table 3: Mean absolute percentage error (%) of the MMD estimates between domains, based on samples drawn by different sampling strategies.

|  | MAPE by held-out domain (%) | | | | |
| --- | --- | --- | --- | --- | --- |
| Sampler | 1 | 2 | 3 | 4 | Average |
| Random | 50.3 ± 2.1 | **20.0 ± 0.5** | 33.9 ± 1.8 | 28.4 ± 1.3 | 33.1 ± 0.8 |
| $k$-DPP | 28.5 ± 2.3 | 26.8 ± 2.2 | **15.9 ± 0.8** | 20.4 ± 2.8 | **22.9 ± 1.1** |
| $k$-means++ | **8.7 ± 0.8** | 55.7 ± 1.1 | 26.5 ± 4.7 | 21.4 ± 4.2 | 28.1 ± 1.6 |

The results show that both diversity-based samplers reduce the MAPE in the small-sample MMD estimates compared to the random sampler, for all but one of the training domain combinations. It is unclear to us why this pattern is reversed for Domain 2; however, the average over all domains is nonetheless favourable. In this case, we can see that the $k$-DPP has produced significantly better MMD estimates than $k$-means++ (despite having higher QE), but, again, this has not directly translated into higher model accuracy.

Overall, these results substantiate our hypothesis that the improved generalisation seen when performing distribution alignment is due to the higher-quality distance estimates generated by diverse samples, but this is of course by no means conclusive proof, and several peculiar phenomena in the results remain to be answered.

## 3. DISCUSSION

This paper introduced a novel use-case of diversity, in the form of enhancing the generalisation power of neural networks trained with distribution alignment. We demonstrated that training on diverse minibatches enabled an adaptive model to surpass the performance of ERM, a result that could not be achieved using standard random sampling methods. Our analysis supported the claim that this was due to the improved distance estimates attained by increasing the distribution coverage of the minibatches.

Regardless of the mechanism by which the performance gain occurs, the notion of a generalised balancing that is not bound by the available labels remains attractive, especially given that performance of the ERM-trained model also improved. It is interesting to note that inducing diversity tends to upweight the importance of outliers in the training set, which is at odds with a common notion in machine learning that outliers should in fact be removed. Specifying relevance or "quality" weights, as was done here for class weights, offers a way to regulate this trade-off. Further exploration of this in the contexts of domain generalisation, distribution alignment, and their application to bioacoustic monitoring, would form a good basis for future work.

## 4. ACKNOWLEDGEMENTS

This work was supported by grants from BAE Systems and the Engineering and Physical Sciences Research Council. The authors acknowledge the use of the IRIDIS High Performance Computing Facility, and associated support services at the University of Southampton, in the completion of this work.